%% file: emnlp2016.tex
\documentclass[11pt,letterpaper]{article}
\usepackage{coling2016}
\usepackage{times}
\usepackage{url}
\usepackage{latexsym}
\usepackage{bm}
\usepackage{graphicx}
\usepackage{amsmath}
\usepackage{amssymb}
\usepackage{harpoon}
\usepackage{CJK}
\usepackage{array}
\usepackage{color}

\input{packages}
\input{MACPdef}

\title{Memory-enhanced Decoder for Neural Machine Translation}

\author{Mingxuan Wang$^1$ \ Zhengdong Lu$^2$ \  
         Hang Li$^2$ \  Qun Liu$^{3,1}$ \\
        $^1$Key Laboratory of Intelligent Information Processing, \\ Institute of Computing Technology, Chinese Academy of Sciences\\
        {\tt \{wangmingxuan,liuqun\}@ict.ac.cn}\\
        $^2$Noah's Ark Lab, Huawei Technologies\\
        {\tt \{Lu.Zhengdong,HangLi.HL\}@huawei.com}\\
        $^3$ADAPT Centre, School of Computing, Dublin City University\\
}

\begin{document}
\maketitle

\begin{abstract}
We propose to enhance the RNN decoder in a neural machine translator (NMT) with external memory, as a natural but powerful extension to the state in the decoding RNN. This memory-enhanced RNN decoder is called \textsc{MemDec}. At each time during decoding, \textsc{MemDec} will read from this memory and write to this memory once, both with content-based addressing. Unlike the unbounded memory in previous work\cite{RNNsearch} to store the representation of source sentence, the memory in \textsc{MemDec} is a matrix with pre-determined size designed to better capture the information important for the decoding process at each time step. Our empirical study on Chinese-English translation shows that it can improve by $4.8$ BLEU upon Groundhog and $5.3$ BLEU upon on Moses, yielding the best performance achieved with the same training set. 
\end{abstract}

\section{Introduction}
The introduction of external memory has greatly expanded the representational capability of neural network-based model on modeling sequences\cite{NTM}, by providing flexible ways of storing and accessing information. More specifically, in neural machine translation, one great improvement came from using an array of vectors to represent the source in a sentence-level memory and dynamically accessing relevant segments of them (“alignment”)  \cite{RNNsearch} through content-based addressing \cite{NTM}. The success of RNNsearch demonstrated the advantage of saving the entire sentence of arbitrary  length in an unbounded memory for operations of next stage (e.g., “decoding”).  

In this paper, we show that an external memory can be used to facilitate the decoding/generation process thorough a memory-enhanced RNN decoder, called \textsc{MemDec}.  The memory in \textsc{MemDec} is a direct extension to the state in the decoding, therefore functionally closer to the memory cell in LSTM\cite{LSTM}.  It takes the form of a matrix with pre-determined size, each column (``a memory cell") can be accessed by the decoding RNN with content-based addressing for both reading and writing during the decoding process. This memory is designed to provide a more flexible way to select, represent and synthesize the information of source sentence and  previously generated words of target relevant to the decoding. This is in contrast to the set of hidden states of the entire source sentence (which can viewed as another form of memory) in \cite{RNNsearch} for attentive read, but can be combined with it to greatly improve the performance of neural machine translator. We apply our model on English-Chinese translation tasks, achieving performance superior to any published results, SMT or NMT, on the same training data ~\cite{xie2011a,DeepMemory,tu2016modeling,tu2015context-dependent}

Our contributions are mainly two-folds
\begin{itemize}
  \item we propose a memory-enhanced decoder for neural machine translator， which naturally extends the RNN with vector state.
  \item our empirical study on Chinese-English translation tasks show the efficacy of the proposed model. 
\end{itemize}

\paragraph{Roadmap} In the remainder of this paper, we will first give a brief introduction to attention-based neural machine translation in Section 2, presented from the view of encoder-decoder, which treats the hidden states of source as an unbounded memory and the attention model as a content-based reading. In Section 3, we will elaborate on the memory-enhanced decoder \textsc{MemDec}. In Section 4, we will apply NMT with \textsc{MemDec} to a Chinese-English task. Then in Section 5 and 6, we will give related work and conclude the paper.

\section{Neural machine translation with attention}
Our work is built on attention-based NMT\cite{RNNsearch},
which represents the source sentence as a sequence of vectors after being processed by RNN or bi-directional RNNs,
and then conducts dynamic alignment and generation of the target sentence with another RNN simultaneously. 

\begin{figure}[h!]
\begin{center}
      \includegraphics[width=0.4\textwidth]{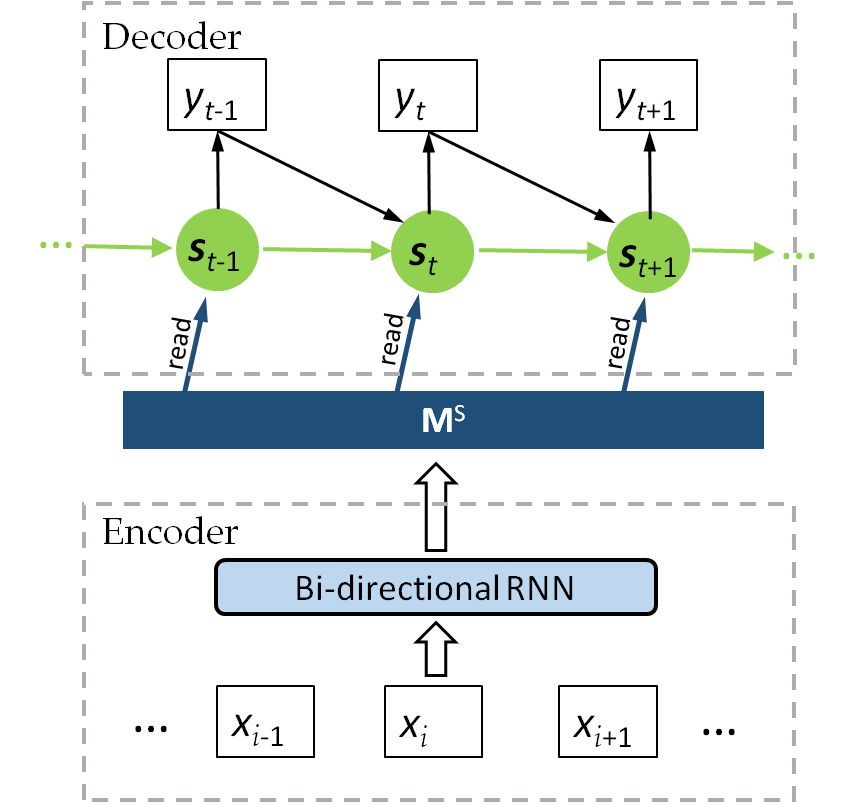}\\
    \caption{RNNsearch in the encoder-decoder view.}
    \label{f:RNNsearch}
  \end{center} 
\end{figure}

Attention-based NMT, with RNNsearch as its most popular representative, generalizes the conventional notion of encoder-decoder in using a unbounded memory for the intermediate representation of source sentence and content-based addressing read in decoding, as illustrated in Figure~\ref{f:RNNsearch}. More specifically, at time step $t$, RNNsearch first get context vector $\c_t$  after reading from the source representation $\M^\textsc{s}$, which is then used to update the state, and generate the word $y_t$ (along with the current hidden state $\sss_t$, and the previously generated word $y_{i-1}$).

Formally, given an input sequence $\mathbf{x}=[x_1, x_2, \dots, x_{T_x}]$
and the previously generated sequence $\mathbf{y}_{<t} = [y_1,y_2,\dots, y_{t-1}]$,
the probability of next word $y_t$ is 
\begin{equation}
  p(y_t|\mathbf{y}_{< t};\mathbf{x})=f(\c_t, y_{t-1},\sss_t),
\end{equation}
where $\sss_t$ is state of decoder RNN  at time step $t$ calculated as
\begin{equation}
  \sss_t = g(\sss_{t-1}, y_{t-1}, \c_{t}).
\end{equation}
where $g(\cdot)$ can be an be any activation function, 
here we adopt a more sophisticated dynamic operator as in Gated Recurrent Unit (GRU, \cite{GRU}). In the remainder of the paper, we will also use \textsf{GRU} to stand for the operator. The reading $\c_t$ is calculated as 
\begin{equation}
  \c_t = \sum_{j=1}^{j=T_x}\alpha_{t,j}\h_j,
\end{equation}
where $\h_j$ is the $j^{th}$ cell in memory $\M^\textsc{s}$. More formally, $\h_j = [\overleftarrow{\h_j}^\top,\overrightarrow{\h_j}^\top]^\top$ is the annotations of $x_j$ and  contains information about the whole input sequence
with a strong focus on the parts surrounding $x_j$, which is computed by 
a bidirectional RNN.
The weight $\alpha_{t,j}$ is computed by
\[
\alpha_{t,j} = \frac{\exp(e_{t,j})}{\sum_{k=1}^{k=T_x}\exp(e_{t,k})}.
\]
where $e_{i,j}=\vv_a^T\tanh(\W_a\sss_{t-1}+\U_a\h_j)$ scores how well $\sss_{t-1}$ and 
the memory cell $\h_j$ match. This is called automatic alignment~\cite{RNNsearch} or attention model~\cite{luong2015effective}, but it is essentially reading with content-based addressing defined in~\cite{NTM}. With this addressing strategy the decoder can attend to the source representation that is most relevant to the stage of decoding.

\begin{figure*}[t!]
\begin{center}
      \includegraphics[width=0.8\textwidth]{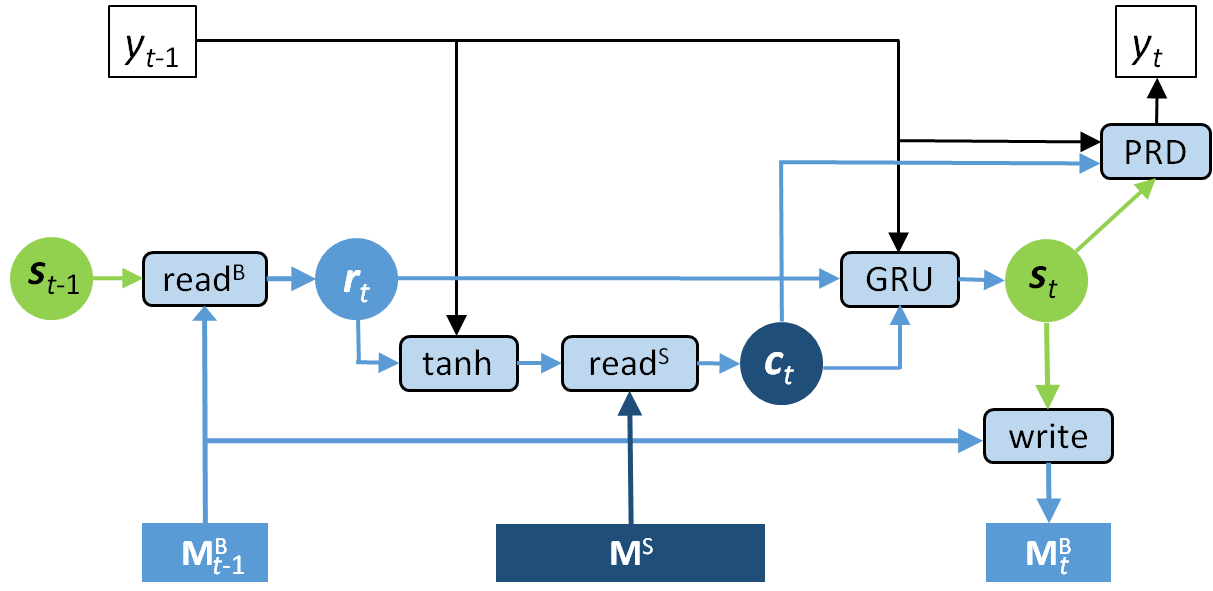}\\
    \caption{Diagram of the proposed decoder \textsc{MemDec} with details.}
    \label{f:decoder-diagram}
  \end{center} 
\end{figure*}

\subsection{Improved Attention Model}
The alignment model $\alpha_{t,j}$ scores how well the output at position $t$ matches the inputs around position $j$ based on $\sss_{t-1}$ and $h_j$.
It is intuitively beneficial to exploit the information of  $y_{t-1}$ when reading from $\M^\textsc{s}$, which is missing from the implementation of attention-based NMT in~\cite{RNNsearch}.
In this work, we build a more effective alignment path by feeding both previous hidden state $\sss_{t-1}$ 
and the context word $y_{t-1}$ to the attention model, inspired by the recent implementation of attention-based NMT\footnote{\url{github.com/nyu-dl/dl4mt-tutorial/tree/master/session2}}.
Formally, the calculation of  $e_{t,j}$ becomes 
\[
e_{t,j}=\vv_a^T\tanh(\W_a\tilde\sss_{t-1}+\U_a\h_j),
\]
where 
\begin{itemize}
  \item $\tilde\sss_{t-1}=\mathcal{H}(\sss_{t-1}, \e_{y_{t-1}})$ is an intermediate state tailored for reading from $\M^\textsc{s}$ with the information of $y_{t-1}$ (its word embedding being $\e_{y_{t-1}}$) added;
  \item $\mathcal{H}$ is a nonlinear function, which can be as simple as $\textsf{tanh}$ or as complex as
\textsf{GRU}. In our preliminary experiments, we found \textsf{GRU} works slightly better than \textsf{tanh} function, but we chose the latter for simplicity.
\end{itemize}


\section{Decoder with External Memory}
\begin{figure}[!h]
\begin{center}
      \includegraphics[width=0.4\textwidth]{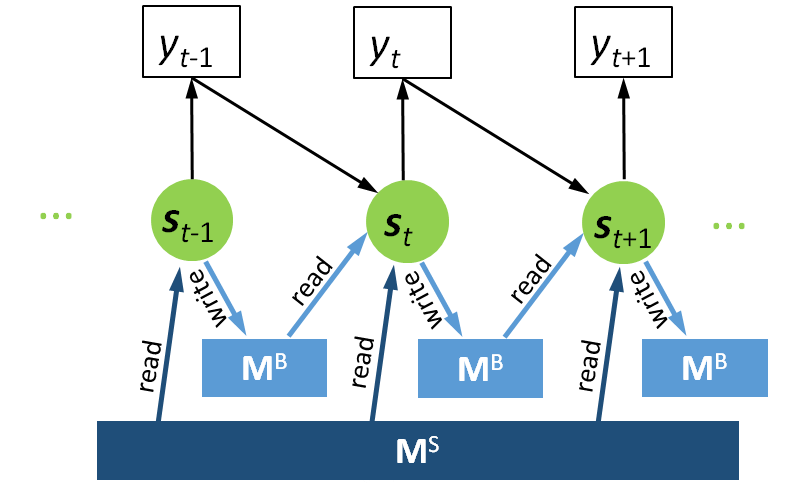}\\
    \caption{High level digram of \textsc{MemDec}.}
    \label{f:MemDec}
  \end{center} 
\end{figure}

In this section we will elaborate on the proposed memory-enhanced decoder \textsc{MemDec}.
In addition to the source memory $\M^\textsc{s}$, \textsc{MemDec} is equipped with a buffer memory $\M^\textsc{b}$ as an extension to the conventional state vector. Figure~\ref{f:MemDec} contrasts \textsc{MemDec} with the decoder in RNNsearch (Figure~\ref{f:RNNsearch}) on a high level.

In the remainder of the paper, we will refer to the conventional state as vector-state (denoted $\sss_t$) and its memory extension as memory-state 
(denoted as $\M^\textsc{b}_t$). Both states are updated at each time step in a interweaving fashion, while the output symbol $y_t$ is predicted based solely on vector-state $\sss_t$ (along with $\c_t$ and $y_{t-1}$). The diagram of this memory-enhanced decoder is given in Figure \ref{f:decoder-diagram}.

\begin{figure}[!h]
\begin{center}
      \includegraphics[width=0.45\textwidth]{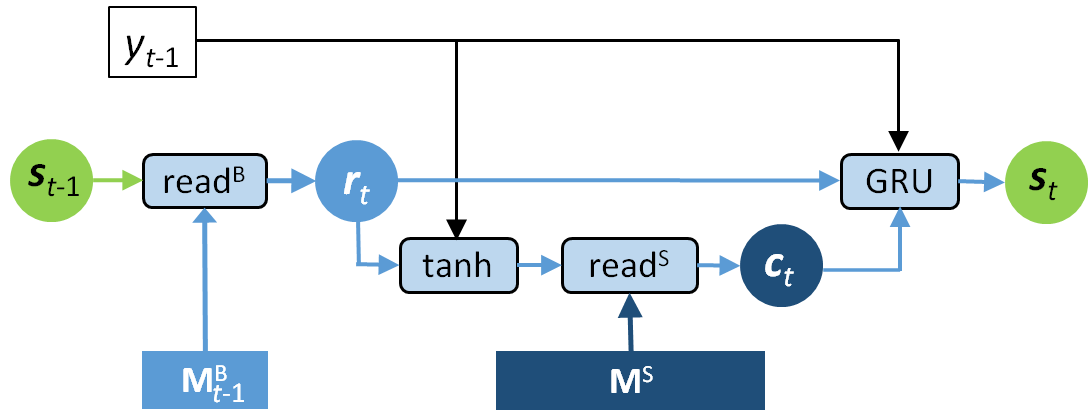}\\
    \caption{Vector-state update at time $t$.}
    \label{f:state-update}
  \end{center} 
\end{figure}

\paragraph{Vector-State Update} 
At time $t$, the vector-state $\sss_t$ is first used to read $\M^\textsc{b}$
\begin{eqnarray}
\rr_{t-1} &=& \textsf{read}^\textsc{b}(\sss_{t-1}, \M^\textsc{b}_{t-1})
\label{e:readbuffer}
\end{eqnarray}
which then meets the previous prediction $y_{t-1}$ to form an ``intermediate" state-vector
\begin{eqnarray}
\widetilde{\sss}_t = \textsf{tanh}(\W^r \rr_{t-1}+\W^y \e_{y_{t-1}}).
\end{eqnarray}
where $\e_{y_{t-1}}$ is the word-embedding associated with the previous prediction $y_{t-1}$.
This pre-state $\widetilde{\sss}_t$ is used to read the source memory $\M^\textsc{s}$
\begin{eqnarray}
\c_{t} &=& \textsf{read}^\textsc{s}(\widetilde{\sss}_t, \M^\textsc{s}).\label{e:readsource}
\end{eqnarray}
Both readings in Eq. (\ref{e:readbuffer}) \& (\ref{e:readsource}) follow content-based addressing\cite{NTM} (details later in Section \ref{s:readingbuffer}). After that, $\rr_{t-1}$ is combined with output symbol $y_{t-1}$ and $\c_t$ to update the new vector-state
\begin{equation}
\sss_{t} = \textsf{GRU}(\rr_{t-1}, \y_{t-1}, \c_{t})
\end{equation}
The update of vector-state is illustrated in Figure~\ref{f:state-update}.

\begin{figure}[!h]
\begin{center}
      \includegraphics[width=0.3\textwidth]{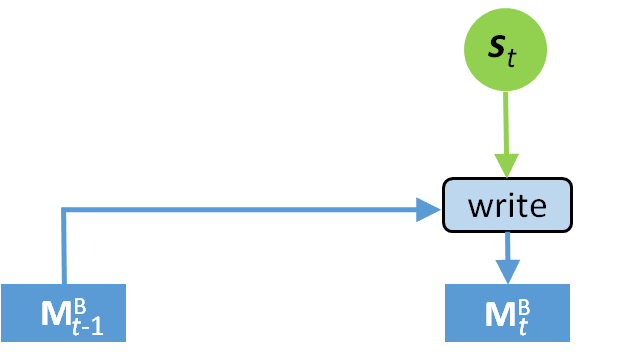}\\
    \caption{Memory-state update at time $t$.}
    \label{f:memory-update}
  \end{center} 
\end{figure}

\paragraph{Memory-State Update}
As illustrated in Figure~\ref{f:memory-update}, the update for memory-state is simple after the update of vector-state: with the vector-state $\sss_{t+1}$  the updated memory-state will be 
\begin{equation}
\M^\textsc{b}_{t} = \textsf{write}(\sss_{t}, \M^\textsc{b}_{t-1}) 
\end{equation}
The writing to the memory-state is also content-based, with same forgetting mechanism suggested in \cite{NTM}, which we will elaborate with more details later in this section.

\begin{figure}[!h]
\begin{center}
      \includegraphics[width=0.2\textwidth]{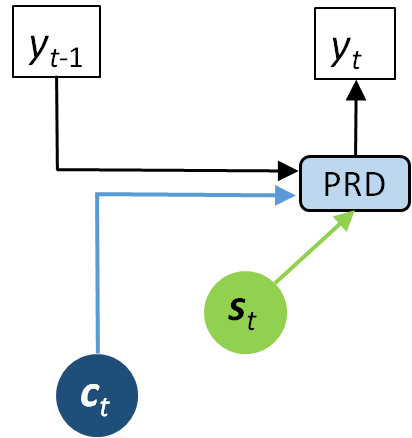}\\
    \caption{Prediction at time $t$.}
    \label{f:prediction}
  \end{center} 
\end{figure}

\paragraph{Prediction}
As illustrated in Figure~\ref{f:prediction}, the prediction model is same as in \cite{RNNsearch}, where the score for word $y$ is given by
\begin{equation}
\textsf{score}(y) =  \textsf{DNN}([\sss_t, \c_t, \e_{y_{t-1}}])^\top \omega_{y}
\end{equation}
where $\omega_{y}$ is the parameters associated with the word $y$. The probability of generating word $y$ at time $t$ is then given by a softmax over the scores
\[
p(y|\sss_t, \c_t, y_{t-1}) = \frac{\exp(\textsf{score}(y))}{\sum_{y'} \exp(\textsf{score}(y'))}.
\]

\subsection{Reading Memory-State} \label{s:readingbuffer}
Formally $\M^\textsc{b}_{t'} \in \mathbb{R}^{n\times m}$ is the memory-state at time $t'$ after the memory-state update,
where $n$ is the number of memory cells and $m$ is the dimension of vector in each cell.
Before the vector-state update at time $t$, the output of reading $\rr_{t}$ is given by 
\[
\rr_{t} = \sum_{j=1}^{j=n} \w^\textsc{r}_{t}(j)\M^\textsc{b}_{t-1}(j)
\]
where $\w^\textsc{r}_{t}\in \mathbb{R}^{n}$ specifies the \emph{normalized} weights assigned to the cells in $\M^\textsc{b}_t$.  Similar with the reading from $\M^\textsc{s}$ ( a.k.a. attention model), we use content-based addressing in determining $\w^\textsc{r}_t$. More specifically, $\w^\textsc{r}_t$ is also updated from the one from previous time $\w^\textsc{r}_{t-1}$ as  
\begin{equation}
\label{e:wr-all}
\w^\textsc{r}_t = g^\textsc{r}_t \w^\textsc{r}_{t-1}+(1-g^\textsc{r}_t)\widetilde{\w}^\textsc{r}_t, 
\end{equation}
where 
\begin{itemize}
  \item $g^\textsc{r}_t = \sigma(\w^\textsc{r}_g\sss_t)$ is the gate function, with parameters $\w^\textsc{r}_g \in \mathbb{R}^{m}$;
  \item  $\widetilde{\w}_{t}$ gives the contribution based on the current vector-state $\sss_t$ 
\begin{eqnarray} \label{e:readweight1}
\hspace{-35pt}\widetilde{\w}^\textsc{r}_{t} \hspace{-4pt}&=& \hspace{-4pt}\text{softmax}(\aa^\textsc{r}_t)\\
\hspace{-35pt}\aa^\textsc{r}_t(i)\hspace{-4pt}&=& \hspace{-4pt} \vv^\top(\W^\textsc{r}_a \M^\textsc{b}_{t-1}(i)+\U^\textsc{r}_a \sss_{t-1}),
\label{e:readweight2}
\end{eqnarray}
with parameters $\W^\textsc{r}_a,\U^\textsc{r}_a \in \mathbb{R}^{m\times m}$ and $\vv \in \mathbb{R}^{m}$.
\end{itemize}

\subsection{Writing to Memory-State}
There are two types of operation on writing to memory-state: \textsc{erase} and \textsc{add}. Erasion is similar to the forget gate in LSTM or GRU,  which determines the content to be remove from memory cells. More specifically, the vector $\mu^\textsc{ers}_t \in \mathbb{R}^m$ specifies the values to be removed on each dimension in memory cells, which is than assigned to each cell through normalized weights $\w^\textsc{w}_t$. Formally, the memory-state after \textsc{erase} is given by
\begin{equation}\label{e:writing}
  \widetilde{\M}^\textsc{b}_t(i) = \M^\textsc{b}_{t-1}(i)(1- \w^\textsc{w}_t(i)\cdot\mu^\textsc{ers}_t)
\end{equation}
\hspace{100pt}$i = 1,\cdots, n$\\
\noindent where 
\begin{itemize}
  \item $\mu^\textsc{ers}_t = \sigma(\W^\textsc{ers}\sss_t)$ is parametrized with $\W^\textsc{ers} \in \mathbb{R}^{m\times m}$;
  \item $\w^\textsc{w}_t(i)$ specifies the weight associated with the $i^{th}$ cell in the same parametric form as in Eq.~(\ref{e:wr-all})-(\ref{e:readweight2}) with generally different parameters.
\end{itemize}
\textsc{add} operation is similar with the update gate in LSTM or GRU, deciding how much 
current information should be written to the memory.
\begin{eqnarray*}
  \M^\textsc{b}_t(i) &=& \widetilde{\M}^\textsc{b}_t(i)+\w^\textsc{w}_t(i)\mu^\textsc{add}_t\\
    \mu^\textsc{add}_t &=& \sigma(\W^\textsc{add} \sss_t)
\end{eqnarray*}
where $\mu^\textsc{add}_t \in \mathbb{R}^m$ and $\W^\textsc{add} \in \mathbb{R}^{m\times m}$.

In our experiments, we have a peculiar but interesting observation: it is often beneficial to use the same weights for both reading (i.e., $\w^\textsc{r}_t$ in Section 3.1) and writing (i.e., $\w^\textsc{w}_t$ in Section 3.2 )  for the same vector-state $\sss_t$. We conjecture that this acts like a regularization mechanism to encourage the content of reading and writing to be similar to each other.  

\subsection{Some Analysis}
The writing operation in Eq. (\ref{e:writing}) at time $t$ can be viewed as an nonlinear way to combine the previous memory-state $\M^\textsc{b}_{t-1}$ and the newly updated vector-state $\sss_t$, where the nonlinearity comes from both the content-based addressing and the gating. This is in a way similar to the update of states in regular RNN, while we conjecture that the addressing strategy in \textsc{MemDec} makes it easier to selectively change some content updated (e.g., the relatively short-term content) while keeping other content less modified (e.g., the relatively long-term content). 

The reading operation in Eq. (\ref{e:wr-all}) can ``extract" the content from $\M^\textsc{b}_t$ relevant to the alignment (reading from $\M^\textsc{s}$) and prediction task at time $t$. This is in contrast with the regular RNN decoder， including its gated variants, which takes the entire state vector to for this purpose. As one advantage, although only part of the information in $\M^\textsc{b}_t$ is used at $t$, the entire memory-state, which may store other information useful for later,  will be carry over to time $t+1$ for memory-state update (writing).

%
%
%
%

\section{Experiments on Chinese-English Translation}
We test the memory-enhanced decoder to task of Chinese-to-English 
translation, where \textsc{MemDec} is put on the top of encoder same as in~\cite{RNNsearch}.

\subsection{Datasets and Evaluation metrics}
Our training data for the translation task consists of $1.25$M sentence
pairs extracted from LDC corpora\footnote{The corpora include LDC2002E18, LDC2003E07, LDC2003E14, Hansards portion of LDC2004T07, LDC2004T08 and LDC2005T06.}, with $27.9$M Chinese words and $34.5$M English words respectively.
We choose NIST 2002 (MT02) dataset as our development set, and the NIST 2003 (MT03),
2004 (MT04) 2005 (MT05) and 2006 (MT06) datasets as our test sets. We use the case-insensitive 4-gram NIST BLEU score
as our evaluation metric as our evaluation metric ~\cite{papineni2002bleu}.

\subsection{Experiment settings}
\paragraph{Hyper parameters}
In training of the
neural networks, we limit the source and target vocabularies to the most frequent $30$K words in both Chinese and
English, covering approximately $97.7\%$ and $99.3\%$ of the two corpora respectively.
The dimensions of word embedding is $512$ and the size of the hidden
layer is $1024$.
The dimemsion of each cell in $\M^\textsc{B}$ is set to $1024$ and the number of cells $n$ is set to $8$. 

\paragraph{Training details}
We initialize the recurrent weight
matrices as random orthogonal matrices. All the
bias vectors were initialize to zero. For other parameters, we initialize them by sampling each element from the Gaussian distribution of mean $0$
and variance $0.01^2$.
Parameter optimization is performed using stochastic gradient descent.
Adadelta ~\cite{zeiler2012adadelta} is used to automatically
adapt the learning rate of each parameter ($\epsilon=10^{-6}$
and $\rho=0.95$). 
To avoid gradients explosion, the gradients of the cost function which had $\ell_2$ norm larger than a predefined threshold $1.0$ was normalized to the threshold ~\cite{pascanu2013construct}.
Each SGD is of a mini-batch of 80 sentences.
We train our NMT model with the sentences of
length up to 50 words in training data, while for moses system we use the full training data.

\paragraph{Memory Initialization}
Each memory cell is initialized with the source sentence hidden state computed as
\begin{eqnarray}
  \M^\textsc{b}(i) &=& \m + \nu_i \\
  \m &=& \sigma(\W_\textsc{ini}\sum_{i=0}^{i=T_x}\h_i)/T_x
\end{eqnarray}
where $\W_\textsc{ini} \in \mathbb{R}^{m \times 2\cdot m}$; $\sigma$ is $\tanh$ function.
$\m$ makes a nonlinear transformation of the source sentence information. $\nu_i$ 
is a random vector sampled from $\calN(0,0.1)$.

\paragraph{Dropout}
we also use dropout for our NMT baseline model and \textsc{MemDec} to avoid over-fitting~\cite{hinton2012improving}.
The key idea is to randomly drop units (along with their connections) from the neural
network during training. This prevents units from co-adapting too much.
In the simplest case, each unit is omitted with a fixed probability $p$, namely dropout rate.
In our experiments, dropout was applied only on the output layer and the dropout rate is set to $0.5$.
We also try other strategy such as dropout at word embeddings or RNN hidden states but fail to get further improvements.
\paragraph{Pre-training}
For \textsc{MemDec}, the objective function is a highly non-convex function of the parameters with more complicated landscape than that for decoder without external memory, rendering direct optimization over all the parameters rather difficult.
Inspired by the effort on easing the training of very deep architectures ~\cite{hinton2006reducing}, we propose a simple pre-training strategy：First we train a regular attention-based NMT model without external memory.
Then we use the trained NMT model to initialize the parameters of encoder and parameters of \textsc{MemDec}, except those related to memory-state (i.e., $\{\W_a^\textsc{r}, \U_a^\textsc{r}, \vv, \w_g^\textsc{r}, \W^\textsc{ers}, \W^\textsc{add} \}$).
After that, we fine-tune all the parameters of NMT with \textsc{MemDec} decoder, including the parameters initialized with pre-training and those associated with accessing memory-state.

\subsection{Comparison systems}
We compare our method with three state-of-the-art systems:
\begin{itemize}
  \item \textbf{Moses:} an open source phrase-based translation system \footnote{\url{http://www.statmt.org/moses/}}: with default configuration
    and a 4-gram language model trained on the target portion of training data.
  \item \textbf{RNNSearch:} an attention-based NMT model with default settings. We use the open source system GroundHog as our NMT baseline\footnote{\url{https://github.com/lisa-groundhog/GroundHog}}.
    \item \textbf{Coverage model}: a state-of-the-art variant of attention-based NMT model ~\cite{tu2016modeling} which improves the attention mechanism through modelling a soft coverage on the source representation.
\end{itemize}

\subsection{Results}

\begin{table*}[!h]
\begin{center}
\begin{tabular}{l|cccc|c}
  \hline
  SYSTEM & MT03 & MT04 & MT05 & MT06 &AVE.\\
  \hline
  \hline
Groundhog & $31.92$ & $34.09$ & $31.56$ & $31.12$ & $32.17$ \\
RNNsearch$^\star$ & $33.11$ & $37.11$ & $33.04$ & $32.99$ &  $34.06$ \\
RNNsearch$^\star$ + \text{coverage} & $34.49$ & $38.34$ & $34.91$ & $34.25$ &  $35.49$ \\
    \hline
\textsc{MemDec} & \textbf{36.16} & \textbf{39.81} & \textbf{35.91} & \textbf{35.98} &\textbf{36.95}\\
\hline \hline
    Moses & $31.61$ & $33.48$ & $30.75$ & $30.85$ & $31.67$ \\
  \hline
\end{tabular}

\end{center}
\caption{\label{tab:EXP} Case-insensitive BLEU scores on Chinese-English translation. Moses is the state-of-the-art phrase-based statistical machine translation system. For RNNsearch, we use the open source system Groundhog as our baseline.  The strong baseline, denoted RNNsearch$^\star$, also adopts \textit{feedback attention} and \textit{dropout}. 
The \textit{coverage} model on top of RNNsearch$^\star$ has significantly improved upon its published version (Tu et al., 2016), which achieves the best published result on this training set. 
For \textsc{MemDec} the number of cells is set to $8$.
}
\end{table*}

The main results of different models are given in Table~\ref{tab:EXP}.
Clearly \textsc{MemDec} leads to remarkable 
improvement over Moses (+$5.28$ BLEU) and Groundhog (+$4.78$ BLEU).
The \textit{feedback attention} gains $+1.06$ BLEU score on top of Groundhog on average,
while together with \textit{dropout} adds another $+0.83$ BLEU score, which constitute the $1.89$ BLEU gain of RNNsearch$^\star$ over Groundhog.
Compared to RNNsearch$^\star$ \textsc{MemDec} is $+2.89$ BLEU score higher, showing the modeling power gained from the external memory.  Finally, we also compare \textsc{MemDec} with
the state-of-the-art attention-based NMT with \textsc{coverage} mechanism\cite{tu2016modeling}, which is about $2$ BLEU over than the published result after adding fast attention and dropout. In this comparison \textsc{MemDec} wins with big margin (+$1.46$ BLEU score).

\subsection{Model selection}
\begin{table*}[!ht]
\begin{center}
\begin{tabular}{l|c|cccc|c}
  \hline
  pre-training& $n$ & MT03 & MT04 & MT05 & MT06 & Ave. \\
  \hline
  \hline
N &4 & $35.29$ & $37.36$ & $34.58$ & $33.32$ & $35.11$ \\
Y &4 & $35.39$ & $39.16$ & $35.33$ & $35.02$ & $36.22$ \\
Y &6 & $35.63$ & $39.29$ & $35.61$ & $34.92$ & $36.58$ \\
Y &8 & $36.16$ & $39.81$ & $35.91$ & $35.98$ & $36.95$ \\
Y &10 & $36.46$ & $38.86$ & $34.46$ & $35.00$ & $36.19$ \\
Y &12 & $35.92$ & $39.09$ & $35.31$ & $35.12$ & $36.37$ \\
  \hline
\end{tabular}
\end{center}
\caption{\label{tab:select} \textsc{MemDec} performances of different memory size.}
\end{table*}

Pre-training plays an important role in optimizing the memory model.
As can be seen in Tab.\ref{tab:select}, pre-training improves upon our baseline 
$+1.11$ BLEU score on average, but even without pre-training our model still gains $+1.04$ BLEU score on average.
Our model is rather robust to the memory size: with merely four cells, our model will be over $2$ BLEU higher than RNNsearch$^\star$. This further verifies our conjecture the the external memory is mostly used to store part of the source and history of target sentence.

\begin{table*}[!ht]\small
  \begin{center}
    \begin{tabular}{l|m{.7\textwidth}}
      \hline
      src& \begin{CJK}{UTF8}{gkai}恩达依兹耶说:``签署(2003年11月停火)协定的各方,最迟必须在元月五日以前把战士的驻扎地点安顿完毕 。''
      \end{CJK} \\
      \hline
      ref & ``All \textit{{\color{blue}{parties}}} that signed the \textit{{\color{blue}{(November 2003 ceasefire)}}} accord should finish the cantoning of their fighters by January 5, 2004, at the latest,'' Ndayizeye said. \\
      \hline
      \textsc{MemDec}&UNK said, `` the \textit{{\color{blue}{parties}} {\color{blue}{
      involved in the ceasefire  agreement on November 2003}}} will have to  be completed by January 5, 2004. '' \\
      \hline
      base& ``The signing of the \textbf{{\color{red}{agreement (UNK-fire) agreement in the November 2003 ceasefire}}} must be completed by January 5, 2004. \\
      \hline
      \hline
      src &\begin{CJK}{UTF8}{gkai}代表团成员告诉今日美国报说,布希政府已批准美国代表团预定元月六日至 十日展开的北韩之行。\end{CJK} \\
        \hline
        ref & Members of the delegation told \textit{{\color{blue}{US Today}}} that the Bush administration had \textit{{\color{blue}{approved the US delegation' s visit}}} to North Korea from January 6 to 10. \\
        \hline
        \textsc{MemDec} & The delegation told the \textit{{\color{blue}{US today}}} that the Bush administration has \textit{{\color{blue}{approved the US delegation's visit}}} to north Korea from 6 to 10 january .\\
        \hline
        base & The delegation told the \textbf{\color{red}{US}} that the Bush administration has \textbf{\color{red}{approved the US to begin his visit}} to north Korea from 6 to 10 January. \\
        \hline

    \end{tabular}
  \end{center}  \caption{\label{tab:sample} \textbf{Sample translations}-for each example, we show the source(src), the human translation (ref),the translation from our memory model \textsc{MemDec} and the translation from RNNsearch(equipped with fast attention and dropout).We
  italicise some \textit{\color{blue}{correct}} translation segments and highlight a few \textbf{\color{red}{wrong}} ones in bold.}\vspace{-3pt}

\end{table*}

\subsection{Case study}
We show in Table 5 sample translations from Chinese to English, comparing mainly \textsc{MemDec} and the RNNsearch model for its pre-training.
It is appealing to observe that 
\textsc{MemDec} 
can produce more fluent translation results and better grasp the semantic information of the sentence.
\section{Related Work}
There is a long thread of work aiming to improve the ability of RNN in remembering long sequences, with the long short-term memory RNN (LSTM)~\cite{LSTM} being the most salient examples and GRU~\cite{GRU} being the most recent one. Those works focus on designing the dynamics of the RNN through new dynamic operators and appropriate gating, while still keeping vector form RNN states. \textsc{MemDec}, on top of the gated RNN, explicitly adds matrix-form memory equipped with content-based addressing to the system, hence greatly improving the power of the decoder RNN in representing the information important for the translation task.


\textsc{MemDec} is obviously related to the recent effort on attaching an external memory to neural networks, with two most salient examples being Neural Turing Machine (NTM)~\cite{NTM} and Memory Network~\cite{memnet}. In fact \textsc{MemDec} can be viewed as a special case of NTM, with specifically designed reading (from two different types of memory) and writing mechanism for the translation task. Quite remarkably \textsc{MemDec} is among the rare instances of NTM which  significantly improves upon state-of-the-arts on a real-world NLP task with large training corpus. 

%
%
%

Our work is also related to the recent work on machine reading~\cite{machinereading}, in which the machine reader is equipped with a memory tape, enabling the model to directly read all the previous
hidden state with an attention mechanism.
Different from their work, we use an external bounded memory and make an abstraction of previous information. In~\cite{DeepMemory}, Meng et. al. also proposed a  deep architecture for sequence-to-sequence learning with stacked layers of memory to store the intermediate representations, while our external memory was applied within a sequence.

%


\section{Conclusion}
We propose to enhance the RNN decoder in a neural machine translator (NMT) with external memory. Our empirical study on Chinese-English translation shows that it can significantly improve the performance of NMT. 
\bibliography{emnlp2016}
\bibliographystyle{emnlp2016}

\end{document}

%% file: packages.tex
\usepackage{amsfonts, amssymb, amsmath}
\usepackage{algorithm, algorithmic, amsthm}
\usepackage{graphicx}
\usepackage{color}
\usepackage{hyperref}
\usepackage[top=2.5cm, bottom=3cm, left=3cm, right=3cm]{geometry}

\usepackage{subfigure}

%% file: MACPdef.tex
\newcommand{\M}{\ensuremath{\mathbf{M}}}

\newcommand{\U}{\ensuremath{\mathbf{U}}}

\newcommand{\W}{\ensuremath{\mathbf{W}}}

\renewcommand{\aa}{\ensuremath{\mathbf{a}}}

\renewcommand{\c}{\ensuremath{\mathbf{c}}}

\newcommand{\e}{\ensuremath{\mathbf{e}}}

\newcommand{\h}{\ensuremath{\mathbf{h}}}

\newcommand{\m}{\ensuremath{\mathbf{m}}}

\newcommand{\rr}{\ensuremath{\mathbf{r}}}
\newcommand{\sss}{\ensuremath{\mathbf{s}}}  

\newcommand{\vv}{\ensuremath{\mathbf{v}}}
\newcommand{\w}{\ensuremath{\mathbf{w}}}

\newcommand{\y}{\ensuremath{\mathbf{y}}}

\newcommand{\calN}{\ensuremath{\mathcal{N}}}

{%
\begin{list}{#1}{
\vspace{-\topsep}
\vspace{-\partopsep}
\setlength{\itemindent}{0cm}
\setlength{\rightmargin}{0cm}
\setlength{\listparindent}{0cm}
\settowidth{\labelwidth}{#1}
\setlength{\leftmargin}{\labelwidth}
\addtolength{\leftmargin}{\labelsep}
\setlength{\itemsep}{0cm}
}%
}%
{%
\end{list}
\vspace{-\topsep}
\vspace{-\partopsep}
}

{\begin{enumerate}%
}%
{\end{enumerate}}

